\pdfoutput=1
\documentclass[11pt]{article}
\usepackage[preprint]{acl}
\usepackage{times}
\usepackage{latexsym}
\usepackage[T1]{fontenc}
\usepackage[utf8]{inputenc}
\usepackage{microtype}
\usepackage{inconsolata}
\usepackage{graphicx}
\usepackage{booktabs}
\usepackage{xcolor}
\usepackage{xspace}
\usepackage{pifont}

\newcommand{\affcmu}{\textsuperscript{\ding{170}}}
\newcommand{\afffujitsu}{\textsuperscript{$\diamondsuit$}}

\newcommand{\socialgym}{\textsc{\mbox{Social}Gym}\xspace}

\title{Exploring Multi-Player Social Games as Verifiable Training Grounds for Large Language Models}
\author{
  \textbf{Keyu He}\affcmu \quad
  \textbf{Michael Li}\affcmu \quad
  \textbf{Jiarui Liu}\affcmu \quad
  \textbf{Mingqian Zheng}\affcmu
\\
  \textbf{Koichi Onoue}\afffujitsu \quad
  \textbf{Tongshuang Wu}\affcmu \quad
  \textbf{Maarten Sap}\affcmu
\\
\\
  \affcmu Carnegie Mellon University, Pittsburgh, USA
\\
  \afffujitsu Fujitsu Research of America, Pittsburgh, USA
\\
  \texttt{\{keyuh, ml7, msap2\}@andrew.cmu.edu}
}

\begin{document}

\maketitle

\begin{abstract}
The training of socially intelligent language models often relies on large language model (LLM) judges, which are expensive, noisy, and vulnerable to reward hacking. 
Multi-player social games offer a natural alternative: they require open-ended communication to coordinate with teammates and infer hidden roles, while their rules provide an objective outcome which can serve as a verifiable reward for training.
Using three social games, we explore how much social intelligence Qwen3 models can acquire through supervised fine-tuning (SFT) and reinforcement learning with verifiable rewards (RLVR).
Results show that against a much stronger Gemma-3-27B model, SFT and RLVR raise the win rate of Qwen3-4B from 33\% to 81\% on the three trained games, and from 39\% to 54\% on three additional held-out games.
In contrast, in-context strategies such as prompting improve the win rate of Qwen3-4B by only about 6 pp across the same games we train and evaluate on.
Furthermore, we show that the gains from SFT and RLVR hold across the model sizes 1.7B, 4B, and 32B.
Qualitative analyses of agent behavior show signs of stronger social intelligence; for example, trained agents tend to share information more openly, check each other's claims against public evidence, and coordinate their votes.
More broadly, our results demonstrate that training LLMs on verifiable outcomes in social games can cultivate generalizable social behaviors, which offers a scalable path for building socially intelligent agents.
\end{abstract}

\section{Introduction}

As agents based on large language models (LLMs) have become more powerful, they have increasingly been used in social tasks, from collaborating on behalf of humans \citep{hua2026passive} to simulating human social behavior \citep{park2023generative,zhou2026odyssim}.
These applications require social intelligence: LLM agents must infer what others know \citep{kim2023fantom,zhou2024reallife}, and decide how to coordinate, collaborate, exchange, and compete with each other \citep{zhou2024sotopia}.
These capabilities are hard to train because social interactions are nuanced and often lack objectively verifiable outcomes.
Existing methods therefore rely largely on LLM judges to provide training signals, using either filtered imitation \citep{wang2024sotopiapi}, preference optimization \citep{kong2025sdpo}, or reinforcement learning \citep{yu2025sotopiarl,wang2026socialrl}.
However, LLM judges are expensive, noisy \citep{zheng2023judging}, and can be exploited via reward hacking \citep{wang2024sotopiapi, zhou2026odyssim}.

\begin{figure*}[t]
\centering
\includegraphics[width=\textwidth]{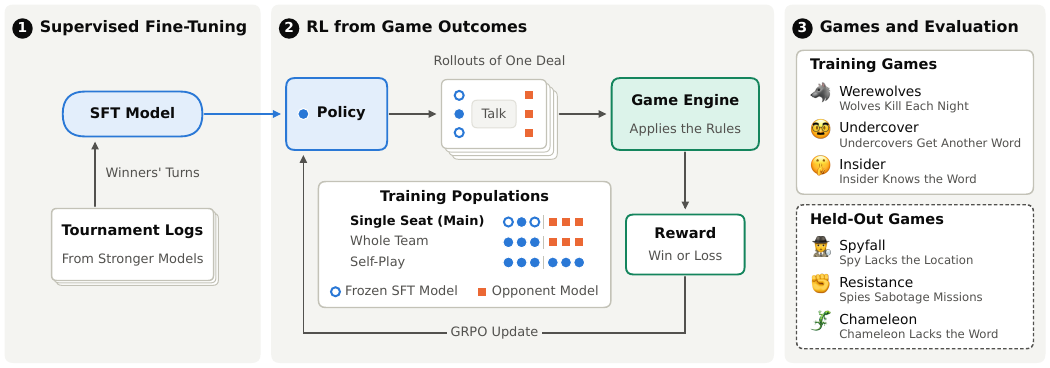}
\caption{Overview of our pipeline. (1) Supervised fine-tuning on winning players' turns from tournament logs. (2) RL from game outcomes; the policy holds one seat beside frozen copies of the SFT model, plays against Gemma-3-27B, and is updated with GRPO from the game outcome. The inset shows the training populations compared in RQ2. (3) The trained policy is evaluated as a whole team on training and held-out games.}
\label{fig:pipeline}
\end{figure*}

Reinforcement learning with verifiable rewards (RLVR) uses rule-based verification to provide objective reward signals and has produced significant gains in LLM reasoning \citep{lambert2025tulu3,guo2025deepseekr1}.
Social games provide such verifiable signals in the context of social interactions, because while interactions may be nuanced, winners are objectively determined by set rules.
This offers a natural extension of RLVR to social contexts.
Prior work has shown that self-play on two-player games can improve LLM reasoning \citep{cheng2024spag,liu2026spiral,yuan2026marshal}.
However, multi-player social games require more complex social reasoning, as agents must coordinate with teammates, infer hidden roles, and persuade players.
In this paper, we introduce a systematic investigation of how multi-agent social games can be used to train language models for improved social gameplay and potentially improved social intelligence in general.
We use \socialgym, an environment containing 22 social games across five categories: normal-form, economic, bluffing, hidden-role deduction, and social strategy \citep{he2026socialgym}.
We specifically ask:
\begin{itemize}
    \item \textbf{RQ1:} Does training on game outcomes improve how language models play multi-player social games, both on the games they train on and beyond them, across held-out games and unseen opponents?
    \item \textbf{RQ2:} Does performance improvement depend on which seats the policy controls during training: a single seat, its whole team, or both teams?
    \item \textbf{RQ3:} How does the social behavior of LLM agents change over successive learning steps?
\end{itemize}

To explore these questions, we train Qwen3-4B \citep{yang2025qwen3} using group relative policy optimization (GRPO) \citep{shao2024deepseekmath} on three hidden-role games from \socialgym. We use the objective game outcome, or whether the game was won or lost, as the verifiable reward. We then evaluate against Gemma-3-27B \citep{gemma2025gemma3}, a model with a base version much stronger than Qwen3-4B on \socialgym games, on both training and held-out games. An overview of our model training and evaluation pipeline is depicted in Figure~\ref{fig:pipeline}.

Our results show that RLVR substantially improves performance on trained games, raising Qwen3-4B's win rate from 57\% after SFT to 81\% after RLVR, while yielding modest gains on held-out games (40\% to 54\%) and against unseen opponents (46\% to 57\% against Qwen3.8-27B).
These improvements hold across model sizes on the \socialgym leaderboard, where trained Qwen3-1.7B and Qwen3-4B match the overall Elo of GPT-4o, and trained Qwen3-32B ranks second only to GPT-5-mini.
Across training configurations, single-seat training provides the most stable gains compared to whole-team training or self-play.
Behaviorally, trained agents learn to share private information more openly, check claims against public evidence, and coordinate votes with teammates.
However, training for too long causes overfitting, often leading to formulaic messages and rigid habits that degrade held-out performance.
Taken together, our findings demonstrate that verifiable outcomes in multi-player games can cultivate transferable social reasoning without relying on expensive or noisy LLM judges, pointing toward diverse, regularized social environments as a scalable path to generalizable LLM social intelligence.

\section{Related Work}

\subsection{Training Social Intelligence}

Most methods for training social agents learn from an LLM judge, often utilizing the SOTOPIA-EVAL \citep{zhou2024sotopia} framework, which rates how well each agent achieves its goal in interactive role-play scenarios.
SOTOPIA-$\pi$ \citep{wang2024sotopiapi} imitates interactions filtered by LLM ratings, SDPO \citep{kong2025sdpo} applies preference optimization to dialogue segments, and Sotopia-RL \citep{yu2025sotopiarl} and SocialRL \citep{wang2026socialrl} run RL on turn-level rewards derived from an LLM judge.
OMAR \citep{jiang2026omar} uses a single model to play every role of a conversation in self-play, including in the multi-player game Werewolf.
\citet{hua2026passive} train a separate Qwen3-4B policy with RL in each of six two-party negotiation and coordination domains, such as price bargaining and meeting scheduling, and find that training in one domain helps another most when the two have a similar interaction structure.
In contrast, our reward is a single score per game computed by the rules, with no LLM judge anywhere in training.
Unlike prior work, we train on several multi-player, team-based social games and test whether the gains carry over to held-out games.

\subsection{RL in Language Games}

Self-play RL on games with rule-decided outcomes can strengthen LLM reasoning, as shown in adversarial taboo with SPAG \citep{cheng2024spag}, zero-sum TextArena games \citep{guertler2025textarena} like Kuhn Poker with SPIRAL \citep{liu2026spiral}, and cooperative or competitive games with MARSHAL \citep{yuan2026marshal}, which also improves held-out play.
Yet all of these settings involve only two players and no hidden teams. Multi-player social deduction games, widely used to evaluate the social capabilities of LLM agents \citep{xu2023exploring,light2023avalonbench,lan2024avalon,bailis2024werewolf}, are now increasingly used as training environments.
Existing training methods for multi-player language games typically add game-specific structure or rewards for particular behaviors.
Cicero \citep{meta2022cicero} combines a dialogue model with strategic planning in Diplomacy.
For Werewolf, \citet{xu2024language} train an RL policy that selects among LLM-generated actions, LSPO \citep{xu2025lspo} alternates game-theoretic optimization in a latent strategy space with preference fine-tuning, and MaKTO \citep{ye2025makto} applies KTO to responses collected from multi-agent play.
\citet{sarkar2025training} reward messages by their influence on other agents in a game based on Among Us, and \citet{zheng2026stackelberg} reward utterances for their persuasive impact.
Most of these methods also train within a single game.
Closest to us, \citet{zhang2026playing} fine-tune LLMs with RL on hidden-role games and find that terminal win-loss rewards alone do not reliably teach social deduction.
We ask what such rewards do teach, and whether those gains transfer to games never seen in training.

\subsection{Generalization from RL Training}

Whether RL training on outcome rewards motivates the learning of generalizable reasoning strategies or narrow shortcuts is a central question.
Compared with pure SFT, RL with outcome rewards generalizes better to unseen task variants \citep{chu2025sft} and across domains \citep{huan2026math}, and it forgets less of what the base model knows \citep{shenfeld2026razor}. However, it remains debated whether RLVR teaches new abilities or merely reweights existing ones \citep{yue2025does,liu2025prorl}.
For social reasoning specifically, \citet{sarangi2026can} find that RLVR on theory-of-mind tasks improves small models in distribution but fails to carry over when the task structure changes.
Multi-player games introduce a second dimension of generalization beyond task structure, as other players are not static.
A policy trained on game outcomes could easily memorize conventions specific to its training games or exploit the habits of a fixed training opponent rather than learning transferable social skills.
To distinguish genuine social reasoning from narrow overfitting, we therefore evaluate transfer across both held-out games and unseen opponents.

\section{Experimental Setup}

\subsection{Terminology}

We define terms used in the rest of this paper. A \emph{game} is a \socialgym environment, such as Werewolves or Spyfall. A \emph{deal} is a starting configuration of roles and models. An \emph{episode} or \emph{rollout} is a complete playthrough of a deal from start to terminal win or loss. A \emph{turn} is a single response or action from a player in an episode.

\subsection{Games}

We train language models using hidden-role team games from \socialgym \citep{he2026socialgym}.
We choose hidden-role games because they best separate models on the leaderboard, and they effectively test the ability to deduce and persuade.
We train on Werewolves, Undercover, and Insider, and hold out Spyfall, Resistance, and Chameleon for evaluation.
Rules for these games are in Appendix~\ref{app:rules}.
In each game, four to six players form two teams: a \emph{majority side} and a \emph{hidden side}, with all identities concealed.
Players receive role-specific information and, on each turn, speak, vote, or take a game-specific action. 
An episode ends when the game's rules declare one side the winner, and this win or loss outcome, derived by the game engine, is the signal that both of our training stages learn from.

\subsection{Supervised Fine-Tuning}

We primarily fine-tune Qwen3-4B-Instruct-2507 \citep{yang2025qwen3}, but also use Qwen3-1.7B and Qwen3-32B to test generalization across model sizes. To teach each model to produce messages and actions in the required format, we first fine-tune with LoRA \citep{hu2022lora} on turns played by the two highest-rated models on the \socialgym leaderboard, GPT-5-mini and GPT-4o, and the highest-rated open-weight model, Gemma-3-27B. For each training game, each of these models plays 54 episodes: three on each of two sides, against each of the nine other models. We only train on turns from won episodes; this ranges from 24 to 45 episodes for each model and game combination, with 292 episodes in total across all three models and all three games.

\subsection{Reinforcement Learning}

We use group relative policy optimization (GRPO) \citep{shao2024deepseekmath} adapted to multi-turn games.
In our main \emph{single-seat} setting, the trainable policy controls one seat, frozen copies of the SFT model fill its teammates' seats, and Gemma-3-27B \citep{gemma2025gemma3} plays the opposing team.
At each update, we sample 36 deals (12 per training game, fixing seating and the environment seed) and play each deal eight times with independent generation samples served via vLLM \citep{kwon2023vllm}.
On each turn, the policy outputs a JSON action following the game's schema; unparseable outputs default to no action with a reward penalty, and the engine assigns positive or negative terminal rewards for wins or losses.
We standardize rewards across the eight rollouts of each deal, assign the resulting advantage to all trainable turns in each rollout, and update the policy with AdamW (learning rate $3\times10^{-5}$) by minimizing the advantage-weighted negative log-likelihood over completion tokens plus a KL penalty (coefficient $0.02$) to the frozen SFT policy.
We report the main checkpoint at 180 updates and train to 240 to examine how longer training affects held-out performance.
To test seat control (RQ2), we also evaluate \emph{whole-team training}, where the policy controls all seats on one team against Gemma-3-27B, and \emph{self-play}, where one policy controls both teams with separate team-level advantages.

\subsection{Evaluation}

We evaluate policies along two axes: controlled head-to-head play and multi-game tournament strength.
For \emph{head-to-head play}, we construct 720 fixed deals (60 per side across the three training and three held-out games, with fixed seating and environment seeds) and test paired win-rate differences with exact McNemar tests.
At test time, the policy controls an entire team rather than a single seat, testing standalone team play against both the training opponent (Gemma-3-27B) and an unseen opponent (Qwen3.8-27B).
For Qwen3-4B, we compare the untrained model, the same model with a GPT-5-generated SPaRTan playbook \citep{he2026socialgym}, the SFT checkpoint, and RLVR checkpoints (main run at 180 and 240 steps, a rerun, and ten lighter runs); for Qwen3-1.7B and Qwen3-32B, we compare SFT and RLVR.
For \emph{tournament strength}, we test transfer beyond hidden-role games by evaluating the SFT and RLVR checkpoints of all three sizes on 17 competitive \socialgym games\footnote{We leave out Bohnanza, the one remaining competitive game on the leaderboard, because its long episodes make it by far the most expensive game to play against API-served models.} across all five categories against nine ranked models, estimating overall and per-category Elo (with 95\% intervals) against a fixed board alongside GPT-4o and GPT-5-mini as references.

\subsection{Behavior and Transfer Analysis}
\label{sec:analysis}

To understand how play changes during training (RQ3), we combine four measures, primarily on the 720 evaluation episodes:
(1)~\emph{Rule-based log metrics}: we track role-specific actions (e.g., majority votes on hidden players, Resistance fail cards, Chameleon secret-word leaks) and surface degeneration (near-verbatim copying and malformed actions).
(2)~\emph{Verified speech-act labels}: blinded to checkpoint and outcome, Claude Sonnet 5.5 labels the policy's public speeches across twelve social and deceptive categories; following the monitor--verifier split of \citet{li2026epog}, we retain a label only if its supporting quote appears verbatim (Appendix~\ref{app:behavior}).
(3)~\emph{SOTOPIA-EVAL}: GPT-6-Luna scores policy seats across seven social dimensions \citep{zhou2024sotopia}, and we classify the weaknesses cited in its believability rationales.
(4)~\emph{Out-of-game transfer}: we evaluate checkpoints on AI-LieDar \citep{su2025ailiedar} to test whether restraint or deception learned in games carries over to assistant dialogues with incentives to conceal negative information.

\section{Results}

\begin{table}[t]
\centering
\small
\setlength{\tabcolsep}{4pt}
\begin{tabular}{@{}lccc@{}}
\toprule
 & \multicolumn{2}{c}{Win rate vs.\ Gemma (\%)} & Elo \\
Model & Trained & Held-out & 17 games \\
\midrule
GPT-5-mini & 64.4 & 56.1 & 1122 \\
GPT-4o & 58.1 & 57.8 & 1066 \\
\midrule
\multicolumn{4}{@{}l}{\emph{Qwen3-1.7B}} \\
\quad No training & 29.2 & 19.7 & \phantom{0}922 \\
\quad SFT & 48.9 & 32.5 & \phantom{0}976 \\
\quad SFT + RLVR & \textbf{80.8} & \textbf{36.7} & \textbf{1056} \\
\multicolumn{4}{@{}l}{\emph{Qwen3-4B}} \\
\quad No training & 33.3 & 38.6 & \phantom{0}953 \\
\quad SPaRTan by GPT-5 & 41.7 & 42.8 & 1007 \\
\quad SFT & 56.7 & 39.7 & 1017 \\
\quad SFT + RLVR & \textbf{81.1} & \textbf{54.4} & \textbf{1060} \\
\multicolumn{4}{@{}l}{\emph{Qwen3-32B}} \\
\quad No training & 42.2 & 45.8 & 1008 \\
\quad SFT & 61.4 & 48.3 & 1052 \\
\quad SFT + RLVR & \textbf{82.2} & \textbf{55.6} & \textbf{1091} \\
\bottomrule
\end{tabular}
\caption{Win rate against Gemma-3-27B on 360 episodes of the three training games and 360 episodes of the three held-out games, and Elo on 17 competitive social games played against nine other ranked models. The untrained Qwen3 models, GPT-4o, and GPT-5-mini are ranked models themselves, so their Elo is the leaderboard rating. Bold: the higher value of each size.}
\label{tab:main}
\end{table}

\begin{figure}[t]
\centering
\includegraphics[width=\columnwidth]{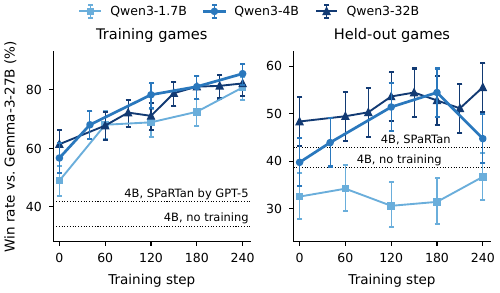}
\caption{Win rate against Gemma-3-27B during RLVR for the three model sizes, each starting from its own SFT model and trained with the same recipe. Bars: 95\% Wilson intervals. Dotted lines: Qwen3-4B without training and with its SPaRTan playbook.}
\label{fig:curves}
\end{figure}

\subsection{RQ1: Training on Game Outcomes Improves Play}

RLVR substantially outperforms both prompting and SFT (Table~\ref{tab:main}).
To separate what prompting, SFT, and RLVR each contribute, we compare the untrained Qwen3-4B, the same model with a SPaRTan playbook, the SFT model, and the RLVR model on the same 720 deals against Gemma-3-27B.
SPaRTan \citep{he2026socialgym}, where the playing model writes its own strategy playbook, is the strongest prompting method from \socialgym and outperforms Reflexion \citep{shinn2023reflexion}, Self-Refine \citep{madaan2023self}, in-context example episodes, and generic game or social advice.
Here, we have GPT-5 write a playbook for each game from records of the untrained Qwen3-4B's episodes, which also beats the other playbook variants and placebo prompts we tried.
On the three training games, the playbook raises Qwen3-4B's win rate from 33.3\% to 41.7\% ($p=0.005$), SFT raises it to 56.7\%, and 180 steps of RLVR reach 81.1\% ($p<0.001$ against the SFT model).
On the three held-out games, however, neither the playbook (42.8\%, $p=0.15$) nor SFT (39.7\%, $p=0.74$) differs significantly from the untrained model (38.6\%), whereas RLVR reaches 54.4\% with $p<0.001$ against SFT.

Training gains are stable across runs, whereas held-out gains depend on the run and length of training. To check that these gains are not specific to one run, we train the main recipe a second time and train ten more runs with a lighter recipe (Appendix~\ref{app:runs}). Every run improves on the training games: the rerun reaches 81.4\% and the ten runs average 75.0\,$\pm$\,2.0\%. On the held-out games, all ten runs end above the SFT model (mean 45.8\,$\pm$\,3.3\%), while the rerun reaches 51.4\% after 60 steps before falling back to 38.1\% at 180 steps. We return to this point when discussing RQ3.

The gains also hold across unseen opponents and model sizes. First, they carry over to an unseen opponent: against Qwen3.8-27B, which never appears in training, the main run wins 57.2\% of the training-game episodes compared with 46.4\% for the SFT model ($p=0.001$), and 37.5\% of the held-out episodes compared with 31.9\% ($p=0.045$). Second, they hold across model sizes (Figure~\ref{fig:curves}): with the same recipe, Qwen3-1.7B and Qwen3-32B gain 32 and 21 pp over their respective SFT models on the training games, and 4 and 7 pp on the held-out games.

Finally, RLVR raises tournament strength beyond hidden-role games (Table~\ref{tab:main}). Ranked across 17 competitive social games against nine other ranked models, Elo rises at every model size: by 80 points for 1.7B (95\% interval 59 to 103), 42 for 4B (18 to 66), and 39 for 32B (15 to 63). This places the trained 1.7B and 4B models level with GPT-4o and the trained 32B model second only to GPT-5-mini. Per-category Elo shows that these gains come from the hidden-role games (122, 71, and 62 points) and, for 1.7B and 32B, from Bargaining, whereas normal-form, bluffing, and social-strategy games do not change significantly. Together, these results show that a rule-decided outcome is a sufficient signal for training social play, and that part of what the policy learns carries over to new games, opponents, and model sizes.

\begin{figure}[t]
\centering
\includegraphics[width=\columnwidth]{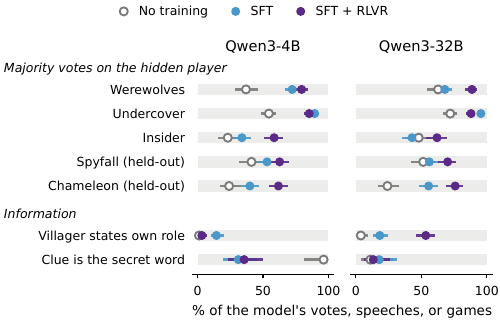}
\caption{Behavior of Qwen3-4B and Qwen3-32B without training, after SFT, and after RLVR, against the nine other ranked models. Top: share of the majority side's votes that land on a hidden player. Bottom: share of the villagers' first-day speeches in Werewolves that state the speaker's own role, and share of Chameleon episodes in which a citizen gives the secret word as its clue. Lines: 95\% Wilson intervals. The untrained models' rates come from leaderboard episodes.}
\label{fig:behavior}
\end{figure}

\subsection{RQ2: Single-Seat Training Yields the Strongest, Most Stable Gains}

Scaling from two-player environments to multi-player environments introduces the design choice of which seats the trainable policy plays during RLVR. To test whether performance gains depend on this choice, we train the same Qwen3-4B SFT model under three settings: (1) single-seat training where one trainable policy plays alongside frozen SFT teammates against Gemma-3-27B, (2) whole-team training where the policy controls all seats on one team against Gemma-3-27B, and (3) self-play where one policy controls both teams. All three settings initially improve play at a similar pace, raising the win rate on the training games from the SFT model's 56.7\% to roughly 68\% within 60 steps: 68.1\% for single-seat training and 68.6\% for self-play after 40 steps, and 67.5\% for whole-team training after 60 steps ($p<0.001$ each). Notably, self-play reaches this level without ever facing Gemma-3-27B, further confirming that outcome-based gains do not require a specific training opponent.

Over longer training, however, single-seat training surpasses both alternatives in held-out generalization and training stability. At its best checkpoint (180 steps), single-seat training wins 81.1\% of the training-game episodes and 54.4\% of the held-out episodes. Whole-team training matches this performance on the training games (80.3\% after 180 steps, $p=0.83$), but its held-out win rate plateaus near 47\% (46.9\% after 60 steps and 46.7\% after 120 steps), falling well short of single-seat training ($p=0.006$ against the 120-step checkpoint). Self-play performs worse: it reaches 42.8\% on the held-out games after 40 steps before collapsing near step 90, where generations run to the token limit even when restarted. By contrast, the single-seat main run, its rerun, and all ten lighter runs complete without collapse. We therefore adopt single-seat training for our main run, showing that verifiable outcome rewards in multi-player games are most effective when a single policy learns alongside frozen teammates against a fixed, stronger opponent.

\begin{table}[t]
\centering
\small
\setlength{\tabcolsep}{3.5pt}
\begin{tabular}{@{}lcccc@{}}
\toprule
 & & & \multicolumn{2}{c}{RLVR steps} \\
\cmidrule(l){4-5}
Win rate (\%) & None & SFT & 180 & 240 \\
\midrule
\multicolumn{5}{@{}l}{\emph{Majority side}} \\
\quad vs.\ Gemma-3-27B & 26.1 & 44.4 & 61.1 & 66.4 \\
\quad vs.\ Qwen3.8-27B (unseen) & 30.8 & 40.8 & 57.2 & 54.2 \\
\multicolumn{5}{@{}l}{\emph{Hidden side}} \\
\quad vs.\ Gemma-3-27B & 45.8 & 51.9 & 74.4 & 63.9 \\
\quad vs.\ Qwen3.8-27B (unseen) & 26.7 & 37.5 & 37.5 & 35.6 \\
\bottomrule
\end{tabular}
\caption{Win rate of Qwen3-4B by side at the four checkpoints of Table~\ref{tab:checkpoints}, pooled over the six games (360 episodes per side). The majority side must find the hidden player; the hidden side must stay hidden. Qwen3.8-27B never appears in training.}
\label{tab:sides}
\end{table}

\begin{table}[t]
\centering
\small
\setlength{\tabcolsep}{3.5pt}
\resizebox{\columnwidth}{!}{%
\begin{tabular}{@{}lrrrr@{}}
\toprule
 & & & \multicolumn{2}{c}{RLVR steps} \\
\cmidrule(l){4-5}
Measure (\%) & None & SFT & 180 & 240 \\
\midrule
\multicolumn{5}{@{}l}{\emph{Majority side}} \\
\quad Votes on hidden player (trained) & 41 & 63 & 82 & 91 \\
\quad Votes on hidden player (held-out) & 38 & 47 & 64 & 65 \\
\quad Speech calls a vote$^{\dagger}$ & 2 & 8 & 32 & 14 \\
\quad Speech cites evidence$^{\dagger}$ & 22 & 14 & 19 & 18 \\
\quad Spyfall: names the location & 38 & 14 & 8 & 3 \\
\multicolumn{5}{@{}l}{\emph{Hidden side}$^{\dagger}$} \\
\quad Frames a known innocent & 17 & 22 & 34 & 40 \\
\quad States a known falsehood$^{\ddagger}$ & 20 & 13 & 8 & 7 \\
\quad White lie & 7 & 10 & 6 & 3 \\
\multicolumn{5}{@{}l}{\emph{Form}} \\
\quad Near-verbatim copy & 19 & 19 & 43 & 24 \\
\quad Incoherent speech$^{\dagger}$ & 17 & 5 & 11 & 9 \\
\quad Believability (0--10)$^{\S}$ & 5.5 & 6.5 & 6.0 & 6.5 \\
\multicolumn{5}{@{}l}{\emph{Hidden roles in held-out games}} \\
\quad Resistance spy plays fail & 82 & 64 & 98 & 60 \\
\quad Caught Chameleon finds word & 61 & 78 & 92 & 37 \\
\bottomrule
\end{tabular}}
\caption{Behavior of Qwen3-4B on the 720 evaluation episodes at four checkpoints of the main run, as rates over the policy's votes, speeches, mission cards, or caught Chameleons. $^{\dagger}$Labeled by Claude Sonnet 5.5. $^{\ddagger}$Without Undercover, whose hidden player copies the Civilians' description (50\% of its speeches at 180 steps). $^{\S}$SOTOPIA-EVAL by GPT-6-Luna.}
\label{tab:checkpoints}
\end{table}

\begin{figure}[t]
\centering
\includegraphics[width=\columnwidth]{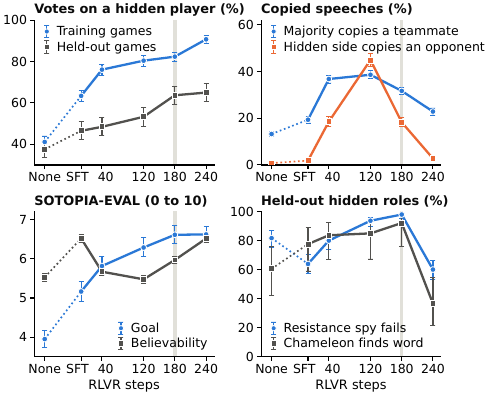}
\caption{Behavior of Qwen3-4B during the main run, on the 720 evaluation episodes. Copied speeches: near-copies of an earlier speech by a teammate (majority side) or an opponent (hidden side). SOTOPIA-EVAL by GPT-6-Luna. Held-out hidden roles: fail cards of Resistance spies and caught Chameleons that name the secret word. Grey line: the main checkpoint. Bars: 95\% Wilson intervals, and bootstrap intervals over episodes for SOTOPIA-EVAL.}
\label{fig:course}
\end{figure}

\subsection{RQ3: RLVR Builds Coordination, Shifts Deception, and Overfits}

Win rates alone do not explain how policies improve. We track the main Qwen3-4B run across four checkpoints on the same 720 deals (untrained, SFT, and RLVR at 180 and 240 steps) using the measures from Section~\ref{sec:analysis} (Table~\ref{tab:checkpoints}). Imitating winning turns via SFT primarily teaches restraint and coherence: Chameleon citizens name the secret word far less often (90\% to 28\% of speeches), and incoherent speech drops from 17\% to 5\%. This restraint extends beyond the games, lowering outright falsification on AI-LieDar \citep{su2025ailiedar} from 62\% to 47\% of dialogues.

\begin{figure}[t]
\centering
\includegraphics[width=\columnwidth]{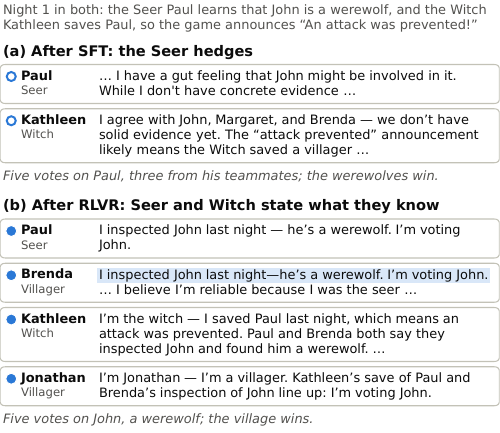}
\caption{Qwen3-32B as the villagers in one Werewolves deal against GPT-5-mini werewolves (John, Margaret). Day-one speeches, shortened; shading marks repeated words.}
\label{fig:paired}
\end{figure}

RLVR then teaches the majority side to coordinate. Across all five voting games, including held-out games, majority players vote for a hidden player more often (Figure~\ref{fig:course}), call votes on named players more frequently (8\% to 32\% of speeches), and cite specific evidence more often in training games (11\% to 17\%) and Spyfall (1\% to 21\%), though Chameleon citizens revert to giving the secret word as their clue (5\% to 20\%). In Werewolves, Seers disclose inspections more often (35\% to 63\%) and unanimous villager voting rises from 35\% to 64\% of rounds; after 180 steps, all villagers coordinate their day-one votes on a werewolf in 37 of 60 episodes (SFT: 15), winning 36 of them (SFT: 9 of 15). These shifts replicate across the ten lighter runs (Appendix~\ref{app:behavior}), and votes on the hidden player rise against the nine other ranked models at both 4B and 32B (Figure~\ref{fig:behavior}), except in Undercover where SFT already votes accurately (about 90\%); at 32B, villagers also begin stating their own role on day one (18\% to 53\%; Figure~\ref{fig:paired}). Crucially, majority-side gains generalize across opponents: against the unseen Qwen3.8-27B, majority win rate climbs from 40.8\% to 57.2\% after 180 steps, matching its gain against Gemma-3-27B (44.4\% to 61.1\%; Table~\ref{tab:sides}).

On the hidden side, RLVR changes how agents deceive rather than how much. Hidden players increasingly frame known innocents (22\% under SFT to 34\% at 180 steps and 40\% at 240 steps), whereas known falsehoods and white lies decline (Table~\ref{tab:checkpoints}). Agents thus favor accusations that the game record cannot refute over refutable factual claims. The same preference carries over to AI-LieDar: outright falsification drops further from 47\% (SFT) to 31\% after 240 steps, while partial lies (concealing or equivocating) rise from 27\% to 45\%, leaving truthful responses statistically unchanged (Appendix~\ref{app:behavior}). Unlike majority-side gains, however, hidden-side gains do not transfer across opponents: win rate stays flat at 37.5\% against Qwen3.8-27B despite jumping from 51.9\% to 74.4\% against Gemma-3-27B (Table~\ref{tab:sides}), suggesting that learned deception adapts to the training opponent.

\begin{figure}[t]
\centering
\includegraphics[width=\columnwidth]{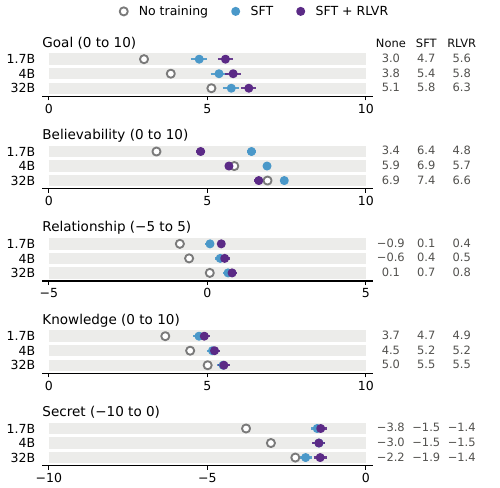}
\caption{SOTOPIA-EVAL scores of the untrained, SFT, and RLVR models at three sizes, in the six hidden-role games against the nine other ranked models, judged by Qwen3.8-27B. Each dimension is shown on its full range; social rules and financial benefits are zero for every model and are omitted. Lines: 95\% bootstrap intervals over episodes. The untrained models' scores come from leaderboard episodes.}
\label{fig:sotopia}
\end{figure}

On SOTOPIA-EVAL, RLVR raises goal and relationship scores while leaving the remaining dimensions largely unchanged, except for believability, which falls. The same tradeoff appears across all three model sizes against the nine other ranked models under Qwen3.8-27B (Figure~\ref{fig:sotopia}), with goal rising by 0.4 to 0.8 points and believability dropping by 0.8 to 1.6 points. Judge rationales (Appendix~\ref{app:behavior}) and rule-based counts trace this drop to temporary speech copying: rationales citing copying jump from 28\% (SFT) to 63\% at 120 steps before receding to 29\% at 240 steps, tracking believability (6.5 to 5.5 and back to 6.5; seats citing copying average 5.2 versus 6.7), while near-verbatim copies follow the same arc (19\% to 52\% to 24\%). Both sides copy strategically (Figure~\ref{fig:course}): majority players echo teammates to align votes (19\% to 39\% at 120 steps), occasionally repeating their own names (Figure~\ref{fig:example}a), whereas hidden players mirror opponents to blend in (2\% to 45\% at 120 steps, dropping to 3\% at 240 steps), including in held-out games (Figure~\ref{fig:example}b). Similarly, the ten lighter runs finish near this copying peak (50\% near-copies), lowering held-out believability (6.01 to 5.16) while raising goal scores (3.97 to 4.45).

Training past 180 steps continues to lift training-game win rates (81.1\% to 85.6\%) but erodes held-out performance (54.4\% to 44.7\%, $p=0.001$; Figure~\ref{fig:curves}, right). The loss concentrates in held-out hidden roles (Figure~\ref{fig:course}): Resistance spies drop from 52 to 27 wins out of 60 ($p<10^{-5}$), reverting from playing fail cards on 98\% of missions to 60\% (near SFT's 64\%), and Chameleon wins fall from 58 to 40 ($p<10^{-3}$), with caught Chameleons identifying the secret word in only 37\% of cases (down from 92\%) amid degenerate speech (such as saying ``I need to eat some cheese'' before guessing their own clue). Meanwhile, malformed actions climb to 8.7\% in held-out games (SFT: 0.6\%; 180 steps: 5.9\%) versus 1.7\% in training games (SFT: 0\%), and training-game speech turns formulaic, with cross-deal template reuse doubling from 9\% (SFT) to 18\% (e.g., ``I'm leaning toward [name].''). Checkpoint selection based on held-out performance is therefore essential. This degradation reflects standard overfitting.

\section{Conclusion}

We asked whether rule-decided outcomes in multi-player social games provide sufficient signal to train language models for social play. Our results show that they do. Training Qwen3-4B with RLVR on three hidden-role games against the much stronger Gemma-3-27B raises its win rate from 57\% after SFT to 81\% on training games and from 40\% to 54\% on held-out games, whereas the strongest prompting baseline, a per-game SPaRTan playbook written by GPT-5, reaches only 42\% and 43\% respectively. The gains hold against an unseen opponent, across 1.7B, 4B, and 32B parameters, and on the 17-game \socialgym leaderboard, where the trained 1.7B and 4B models match the Elo of GPT-4o. Among the training setups, single-seat training, where one trainable policy plays alongside frozen SFT teammates, yields the largest and most stable gains, showing that verifiable outcome rewards do not require self-play. Our behavior analysis shows what the reward teaches: the majority side learns to coordinate its votes, and its gains transfer to the unseen opponent, while the hidden side shifts from known falsehoods to accusations that the game record cannot refute. Checkpoint selection using held-out games is essential, as training for too long leads to overfitting on training games and erosion of performance on held-out games. Together, these findings show that rule-decided game outcomes offer a verifiable, judge-free training signal for social reasoning. Broader game collections and opponent pools are a natural next step toward generalizable social intelligence.

\section*{Limitations}

First, we train on three hidden-role games, and our evidence for transfer comes from three held-out games of the same category and the 17-game \socialgym leaderboard. On that leaderboard, Elo gains come mostly from the hidden-role games and Bargaining, while the normal-form, bluffing, and social-strategy games do not change significantly. Training and evaluating across multiple game categories would test broader social reasoning capabilities. Second, every run trains against a single opponent, Gemma-3-27B. The majority side's gains transfer to an unseen opponent, whereas the hidden side's do not (37.5\% against Qwen3.8-27B both before and after RLVR), suggesting that a diverse opponent pool may be needed to learn transferable deceptive strategies. Third, we train only on the Qwen3 model family, and held-out gains vary across runs and training lengths. We report the main run at its best held-out checkpoint; the ten lighter runs, each evaluated at a fixed step without checkpoint selection, also end above the SFT model on held-out games, though a separate set of validation games would make checkpoint selection fully independent of the reported results. Fourth, part of our behavior analysis relies on LLM judges. We reduce this dependence by retaining only speech-act labels whose supporting quotes appear verbatim, checking judged trends against rule-based log metrics, and using three SOTOPIA-EVAL judges, but we did not run a human evaluation. We also do not have humans participate in our games. Finally, hidden-role games reward deception: our trained agents frame known innocents more often and, on AI-LieDar, shift from outright falsification toward concealing or equivocating. Applying outcome-based RL to social agents outside research should therefore monitor how deceptive behavior shifts in and out of games.

\bibliography{references}

\appendix

\section{Game Rules}
\label{app:rules}
We summarize the rules as the \socialgym engine \citep{he2026socialgym} stated them to the players in our runs. In every game, players speak in a fixed seating order and vote simultaneously.

\paragraph{Werewolves (training).} Two Werewolves (the hidden side), who know each other, play against four villagers: a Seer, a Witch, and two Villagers. Each night, the Werewolves choose a player to kill, the Seer learns whether one player is a Werewolf, and the Witch may save the night's victim or poison a player, each potion only once. Each day, the night's deaths are announced, every living player speaks once, and the player with the most votes is executed. The villagers win when no Werewolf remains, and the Werewolves win when they equal or outnumber the villagers.

\paragraph{Undercover (training).} Four Civilians share a secret word, and two Undercovers (the hidden side) hold a different but related one, such as \emph{vacuum cleaner} against \emph{lawn mower}. Each player sees only their own word. Each round, every remaining player describes their word in one sentence without naming it, and then the player with the most votes is eliminated and their side is announced. The Civilians win when both Undercovers are eliminated, and the Undercovers win when one Undercover and one Civilian remain.

\paragraph{Insider (training).} The Insider (the hidden side) knows the secret word, and four Citizens know only its category. The players take turns asking yes/no questions, which the game answers, up to 15 in total, and then all guess the word at once. If nobody guesses it, all players lose. Otherwise, every player speaks once about who the Insider is and all vote; the Citizens win if the Insider receives the most votes, and the Insider wins otherwise.

\paragraph{Spyfall (held-out).} Three players share a secret location, such as a police station, and the Spy (the hidden side) does not know it. Each round, the remaining players take turns questioning each other, each trying to show that they know the location without giving it away, and then the player with the most votes is eliminated. The other players win when the Spy is eliminated, and the Spy wins when only one other player remains.

\paragraph{Resistance (held-out).} Three Resistance members play against two Spies (the hidden side), who know each other, and nobody is eliminated. Before each proposal, every player speaks once; then a rotating leader proposes a team for the next mission, and all players vote on it. The team goes on the mission if approvals outnumber rejections, and five rejections in a row make the Spies win. On a mission, each team member secretly plays a success or a fail card, Resistance members must play success, and a single fail card fails the mission; only the number of fail cards is announced. The five missions take teams of two, three, two, three, and three players. The Resistance wins once three missions succeed, and the Spies win once three fail.

\paragraph{Chameleon (held-out).} Four Citizens and the Chameleon (the hidden side) see the same topic card, a category with 16 words; the Citizens know which word is secret, and the Chameleon does not. Each player in turn gives a one-word clue, then every player speaks once and all vote, with ties broken by the dealer, who gave the first clue. If the accused player is a Citizen, the Chameleon wins; if it is the Chameleon, it wins by naming the secret word and loses otherwise.

\section{Run-to-Run Variation}
\label{app:runs}

\begin{figure}[t]
\centering
\includegraphics[width=\columnwidth]{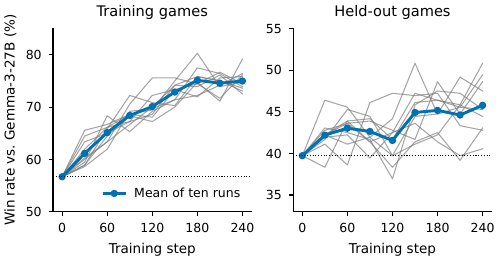}
\caption{Win rate against Gemma-3-27B during training for the ten runs of the lighter recipe (grey) and their mean (blue), on the training games (left) and the held-out games (right). Dotted line: the SFT model.}
\label{fig:tenruns}
\end{figure}

\begin{table}[t]
\centering
\small
\setlength{\tabcolsep}{3.5pt}
\begin{tabular}{@{}rcccc@{}}
\toprule
 & \multicolumn{2}{c}{Training games} & \multicolumn{2}{c}{Held-out games} \\
Step & Mean $\pm$ SD & Range & Mean $\pm$ SD & Range \\
\midrule
0 (SFT) & 56.7 & -- & 39.7 & -- \\
60 & 65.1\,$\pm$\,1.9 & 61.9--68.3 & 43.0\,$\pm$\,2.0 & 38.6--45.6 \\
120 & 70.1\,$\pm$\,2.4 & 67.2--75.6 & 41.6\,$\pm$\,3.0 & 36.9--47.2 \\
180 & 75.1\,$\pm$\,2.5 & 71.9--80.3 & 45.2\,$\pm$\,2.5 & 41.4--48.6 \\
240 & 75.0\,$\pm$\,2.0 & 72.5--79.2 & 45.8\,$\pm$\,3.3 & 40.6--50.8 \\
\bottomrule
\end{tabular}
\caption{Win rate (\%) against Gemma-3-27B of the ten runs at each evaluated step: mean $\pm$ standard deviation over runs and the range from the lowest to the highest run. Step 0 is the SFT model.}
\label{tab:tenruns}
\end{table}

\paragraph{Ten lighter runs.} To measure how much one RLVR recipe varies between runs, we train ten runs of 240 steps from the same SFT model (Figure~\ref{fig:tenruns}, Table~\ref{tab:tenruns}). They use a lighter recipe than the main run: each update samples 18 deals (6 per training game) instead of 36, so a step has half the episodes, and the next step's rollouts are played while the current update runs, with truncated importance weights correcting for the one-update lag. Five runs share the deals and the LoRA initialization and differ only in sampling; the other five use different seeds for both. On the training games, every run improves at every checkpoint, and the runs differ by a standard deviation of 1.9 to 2.5 pp. On the held-out games, the mean rises from the SFT model's 39.7\% to 45.8\% after 240 steps, with a standard deviation of 2.0 to 3.3 pp across runs, and runs that share deals and initialization spread as widely as runs with different seeds.

\section{Behavior Analysis}
\label{app:behavior}

\paragraph{Speech labels.} Claude Sonnet 5.5 read each evaluation episode with every player's role and private information, but without the checkpoint, the episode's identifier, or its result, and labeled each public speech of the policy's team with any of twelve labels: a false claim about the speaker's own role; a game fact it knows to be false, including an Undercover player describing the other word as its own; an accusation of a player it knows to be innocent (framing); a white lie, that is, feigned agreement, praise, thanks, or reassurance that its knowledge or its own next vote contradicts; evading a question or accusation; truthfully disclosing private information; citing a specific vote, clue, answer, mission result, or death as evidence; reasoning about what another player knows or intends (theory of mind); calling a vote; declaring an alliance; repeating earlier speeches without adding anything (echo); and speech inconsistent with the speaker's identity, role, or earlier words (incoherent). The prompt states what each role knows, so a citizen's wrong accusation counts as a mistake, not a lie. Each label carries a quote, and the 68 of 19{,}798 labels whose quote does not appear verbatim in the speech were dropped. Reading 40 random labels, we judged all 16 framing and white-lie labels and 7 of 8 echo labels correct, but only about 6 of 8 evidence and theory-of-mind labels, so we report those two only for large differences.

\paragraph{Why believability falls.} For every policy seat, Claude Sonnet 5.5 listed which of nine weaknesses the SOTOPIA-EVAL judge's believability rationale names (Table~\ref{tab:belreasons}). A second SOTOPIA-EVAL judge, Claude Sonnet, cites copying with the same rise and fall (33\%, 53\%, and 37\% of seats for the SFT model and after 180 and 240 steps).

\begin{table}[t]
\centering
\small
\setlength{\tabcolsep}{3.5pt}
\begin{tabular}{@{}lrrrrr@{}}
\toprule
 & & \multicolumn{4}{c}{RLVR steps} \\
\cmidrule(l){3-6}
 & SFT & 40 & 120 & 180 & 240 \\
\midrule
Believability (0--10) & 6.52 & 5.67 & 5.48 & 5.97 & 6.52 \\
\midrule
\multicolumn{6}{@{}l}{\emph{Weakness named (\% of seats)}} \\
\quad Copies others & 28 & 60 & 63 & 49 & 29 \\
\quad Plays against its role & 33 & 29 & 21 & 23 & 22 \\
\quad Incoherent & 13 & 12 & 11 & 18 & 18 \\
\quad Unnatural style & 3 & 11 & 14 & 10 & 8 \\
\quad Invalid action & 2 & 2 & 12 & 4 & 6 \\
\quad None & 27 & 15 & 14 & 20 & 32 \\
\bottomrule
\end{tabular}
\caption{SOTOPIA-EVAL believability of the policy's 1{,}860 seats per checkpoint (judge GPT-6-Luna) and the weaknesses its rationales name, as classified by Claude Sonnet 5.5.}
\label{tab:belreasons}
\end{table}

\begin{figure*}[t]
\centering
\includegraphics[width=\textwidth]{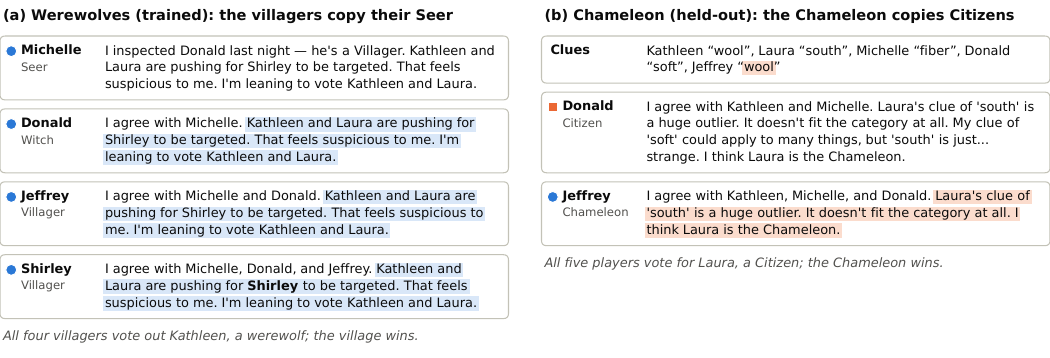}
\caption{Two evaluation episodes after 180 steps, quoted from the logs. Shading marks words repeated from an earlier speech or clue of the same episode. Circles: the policy's players; square: a Gemma-3-27B player. (a)~The policy's villagers repeat their Seer word for word, Shirley included, and vote out a werewolf. (b)~In a held-out game, the policy's Chameleon repeats a Citizen's clue and accusation, and the players vote out an innocent Citizen.}
\label{fig:example}
\end{figure*}

\paragraph{Copying and formulaic speech.} A speech of the policy counts as a near-copy when the sets of its words and of one of the 40 speeches before it in the same episode overlap by at least 60\% (Jaccard), both having at least six words. Its source is the best-matching earlier speech, by a teammate, an opponent, or the speaker itself. A speech reuses a template when, with every player's name replaced by one token, its words overlap by at least 80\% with those of a policy speech for the same side of the same game in another deal. Template reuse in the training games is 9\% for the SFT model, 8\% to 11\% after 40 to 180 steps, and 18\% after 240 steps; in the held-out games it stays below 2\%.

\paragraph{Ten runs.} The rule-based measures on the final checkpoints of the ten lighter runs repeat the main run's shifts in every run: votes on a hidden player in the training games rise from the SFT model's 63\% to between 76\% and 82\%, unanimous villager votes in Werewolves from 35\% to between 57\% and 76\% of rounds, Seer speeches that announce an inspection from 35\% to between 60\% and 100\%, and near-copies from 19\% to between 45\% and 59\% of speeches. Votes on a hidden player in the held-out games rise in nine of the ten runs (47\% to between 44\% and 62\%). Every run also makes Chameleon citizens give away the secret word as their clue more often than the SFT model does (5\% to between 8\% and 37\% of clues).

\paragraph{AI-LieDar.} We ran the exp2 setting of \citet{su2025ailiedar} (290 dialogues), with GPT-6-Luna playing the human and both judges, so only our checkpoints can be compared with each other. The untrained model is truthful in 15.5\% of dialogues, partially lies in 22.4\%, and falsifies in 62.1\%; the SFT model in 26.2\%, 27.2\%, and 46.6\%. Its replies are about half as long (371 against 672 characters) and more often name the problem (``To be transparent, there is a known issue with the zipper on model A''), where the untrained model says ``I can confidently say there are no known issues.'' RLVR lowers falsification further, to 40.1\% of dialogues after 180 steps and 30.8\% after 240 steps, which is 15.9 points below the SFT model on the same dialogues (95\% bootstrap interval 9.3 to 22.5). The difference moves to partial lies (39.1\% and 45.3\%) rather than to truthful answers (20.8\% and 23.9\%, neither significantly different from the SFT model's 26.2\%). The RLVR agent also takes no action on about 4\% of its turns, against 0.1\% for the SFT model; one dialogue per RLVR checkpoint in which the agent never answered could not be judged and is excluded.

\paragraph{Examples.} All from the same deals across checkpoints.
In a Chameleon deal with the secret word \emph{handball}, the untrained citizens give the clues ``handball'' three times and the caught Chameleon wins by guessing it; the SFT citizens give ``net'' and ``court,'' and the Chameleon guesses ``basketball'' and loses; after 180 steps, one citizen's clue is ``team sport where you use your hands to pass and shoot.''
In Werewolves after 180 steps, the Seer opens with ``I inspected Donald last night---he's a Villager. Kathleen and Laura are pushing for Shirley to be targeted. That feels suspicious to me,'' and the other three villagers repeat the rest of the speech word for word, Shirley included; all four vote out Kathleen, a werewolf.
A Resistance spy after 240 steps frames a loyal player: ``Robert's consistent push for himself is very suspicious.''
A caught Chameleon after 180 steps guesses \emph{rabbit} correctly; on the same deal after 240 steps, it says ``I need to eat some cheese. It was a different route, but I had a clue'' and guesses ``moo,'' its own clue.

\end{document}